\documentclass[letterpaper, 10 pt, conference]{ieeeconf}

\IEEEoverridecommandlockouts  
\usepackage{mathtools}
\usepackage{bm} 
\usepackage{amssymb}
\usepackage{amsfonts}
\usepackage{mathbbol} 
\usepackage{graphicx, import} 
\usepackage{float}
\usepackage{times}
\usepackage{multicol}
\usepackage{multirow}    
\usepackage{cases}
\usepackage{yfonts}
\usepackage{subcaption}
\usepackage{colortbl}
\usepackage{xparse}
\usepackage{arydshln}
\usepackage[mediumspace,mediumqspace,squaren]{SIunits}

\usepackage{soul}
\usepackage{dblfloatfix}

\usepackage{tabularx,booktabs}
\newcolumntype{C}{>{\centering\arraybackslash}X} 
\makeatother
\makeatletter
\newsavebox\myboxA
\newsavebox\myboxB
\newlength\mylenA
\newcommand*\xoverline[2][0.75]{%
   \sbox{\myboxA}{$\m@th#2$}%
   \setbox\myboxB\null
   \ht\myboxB=\ht\myboxA%
   \dp\myboxB=\dp\myboxA%
   \wd\myboxB=#1\wd\myboxA
   \sbox\myboxB{$\m@th\overline{\copy\myboxB}$}
   \setlength\mylenA{\the\wd\myboxA}
   \addtolength\mylenA{-\the\wd\myboxB}%
   \ifdim\wd\myboxB<\wd\myboxA%
      \rlap{\hskip 0.5\mylenA\usebox\myboxB}{\usebox\myboxA}%
   \else
       \hskip -0.5\mylenA\rlap{\usebox\myboxA}{\hskip 0.5\mylenA\usebox\myboxB}
   \fi}
\makeatother

\usepackage[x11names]{xcolor}  

\title{\bf Shared Control of Robot-Robot Collaborative Lifting with \\ Agent Postural and Force  Ergonomic Optimization}
\author{Lorenzo Rapetti~$^{1,2}$,
Yeshasvi Tirupachuri~$^{1}$,
Alberto Ranavolo~$^{3}$, \\
Tomohiro Kawakami~$^{4}$,
Takahide Yoshiike~$^{4}$,
and Daniele Pucci~$^{1}$
\thanks{$^{1}$ Dynamic Interaction Control at Istituto Italiano di
 Tecnologia, Center for Robotics Technologies, Genova,
  Italy. ({email: \tt\small name.surname@iit.it})}
\thanks{$^{2}$ Machine Learning and Optimisation, The University of Manchester,
 Manchester, United Kingdom.}
\thanks{$^{3}$  Department of Occupational and Environmental Medicine, Epidemiology
and Hygiene, INAIL, Roma, Italy.}
\thanks{$^{4}$ Honda R\&D Co., Ltd., Saitama, Japan.}
}

\begin{document}

\maketitle
\thispagestyle{empty}
\pagestyle{empty}

\begin{abstract}
Humans show specialized strategies for efficient collaboration.
Transferring similar strategies to humanoid robots can improve their capability to interact with other agents, leading the way to complex collaborative scenarios with multiple agents acting on a shared environment.
In this paper we present a control framework for robot-robot collaborative lifting.
The proposed shared controller takes into account the joint action of both the robots thanks to a centralized controller that communicates with them, and solves the whole-system optimization. Efficient collaboration is ensured by taking into account the ergonomic requirements of the robots through the optimization of posture and contact forces.  
The framework is validated in an experimental scenario with two iCub humanoid robots performing different payload lifting sequences. 
\end{abstract}



\section{Introduction}
\label{section:introduction}
The success of robots in real-world scenarios is still largely dependent on their ability to interact with the environment and with other agents, possibly with a degree of intelligence. Traditionally, robots interact with inanimate objects and take unilateral actions to achieve specific objectives. Today's robots, however, are more frequently asked to collaborate with each other to achieve complex tasks, or to interact with humans either to assist them or to augment their capabilities~\cite{koeppe2005robot}. Collaborative scenarios have thus become a priority for both the industry and the scientific community~\cite{robotics2020}. For successful and efficient completion of future scenarios, it is crucial to implement control strategies that coordinate the interactions of several agents. 
This paper introduces control strategies that enable energetically efficient robot-robot interaction for collaborative lifting of payloads, here referred to as \emph{ergonomic collaborative lifting}.

\begin{figure}[t]
\centering
\begin{subfigure}[b]{0.45\textwidth}
\centering
\includegraphics[width=0.85\columnwidth]{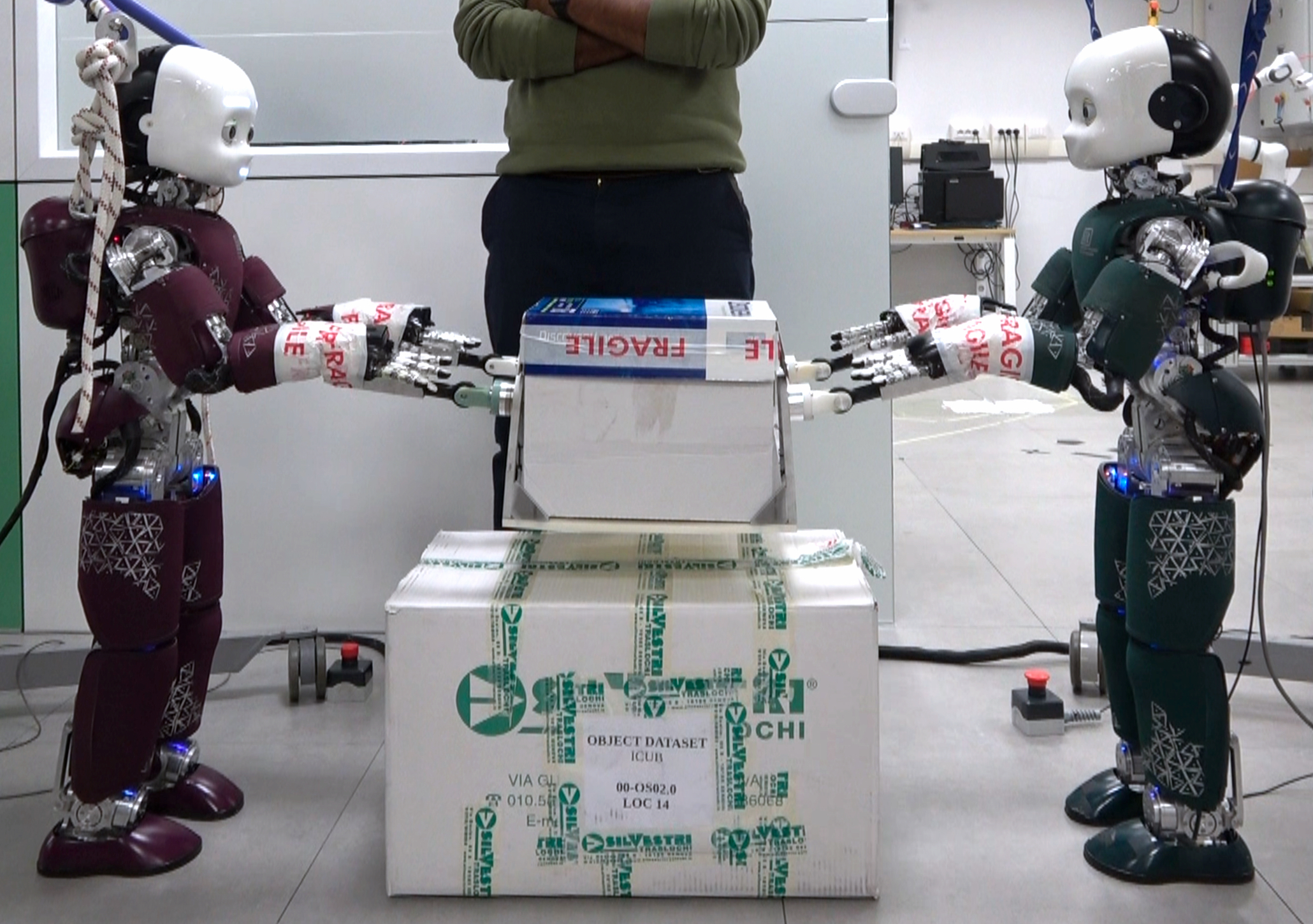}
\end{subfigure}
\caption{Robot-robot collaboration scenario. Two iCub humanoid robots collaborate in lifting a payload.}
\label{fig:real-robots}
\end{figure}

In industrial environments, robots can be used for heavy payloads lifting and carrying activities, relieving human operators from repeated and excessive efforts~\cite{ajoudani2018progress,vysocky2016human}. As a consequence, robots can contribute to make workplaces more ergonomic by reducing the risk factor associated with work-induced musculoskeletal disorders~\cite{wmds}. Our objective is to understand how humanoid robotic platforms can help towards this direction, in particular, by endowing them with collaborative lifting capabilities, e.g. Fig.~\ref{fig:real-robots}. In fact, humans have shown specialized efficient payload sharing strategies~\cite{reed2006}, and
specific control frameworks have been proposed for human-robot collaboration scenarios~\cite{ikeura2002,griffin2005,lawitzky2010,Magrini2015}. The existing strategies, however, 
mainly focus on modelling and stabilizing each system separately, and the advantages arising from joint actions are seldom explored. Hence, centralized approaches for agent-agent interaction can lead to unprecedented applications of robotic systems~\cite{yan2009control}.

Concerning humanoid robot platforms, state-of-the-art whole-body controllers often focus on maintaining robot stability by suppressing external disturbances. Particular attention has been paid to \textit{momentum-based} strategies that have been applied successfully to different platforms \cite{stephens2010dynamic,righetti2011inverse,herzog2014balancing,nava2016stability,kanemoto2018compact}. The main idea behind those strategies consists in controlling the robot momentum through full body joint torques, which are computed via inverse dynamics from desired feet contact forces. For the particular case of multi-contact motion, the contact forces at the hands can also be considered~\cite{audren2014model,kamioka2015dynamic}. These control strategies enable the robot to handle unexpected contacts while being compliant with respect to external interactions~\cite{hyon2007full,pucci2016,ko2018}. A payload lifting task, however, requires the control of interaction forces as well, which should not be treated as disturbances. Additionally, in presence of a second collaborative agent, its actions need to be taken into account as they contribute to the achievement of the task.
Whole-body control strategies enabling real humanoid robots to lift a payload either autonomously~\cite{harada2005,arisumi2008}, or by collaborating with a human partner~\cite{evrard2009,sheng2014,Agravante2019} have been proposed. However, control strategies for efficient humanoid robot-robot collaborative payload sharing and co-manipulation are still missing in the current literature.

This paper presents a novel \textit{momentum-based} whole-body shared control strategy for robot-robot collaboration. The controller achieves balancing of the robots and the control of a payload position-and-orientation during lifting task execution. Additionally, we discuss robot ergonomics in terms of efficient energy consumption, and two concurrent ergonomic metrics are proposed, here referred to as \emph{force ergonomics} and \emph{postural ergonomics}. The proposed approach extends and encompasses the partner-aware approach proposed in~\cite{tirupachuri2019,tirupachuri2019-des}, where the partner actions are exploited in an unidirectional way.
Differently from previous works, we propose a bidirectional optimization of the agents action thanks to a centralized control, and coupled system dynamics modelling. Thanks to modern technologies, in fact, it is possible to solve and share online full body dynamics among the agents \cite{latella2019simultaneous,lorenzini2020online}.
Feasibility and performances of the proposed control approach are verified using two iCub humanoid robots~\cite{Natale2017}.

The paper is organized as follows: Sec.~\ref{section:background} introduces the notation and modelling. Sec.~\ref{section:ergonomy} presents robot ergonomics, used in the proposed shared control strategy presented in Sec.~\ref{section:method}. Sec.~\ref{section:experiments} presents experimental results, and the validation of the proposed approach is discussed. Sec.~\ref{section:conclusioins} concludes the article with conclusions and perspectives.





\section{Background}
\label{section:background}

\subsection{Notation}
\label{sec:background:notation}
\begin{itemize}
    \item $\mathcal{I}$ denotes an inertial frame of reference.
    \item $e_i \in R^m$ is the canonical vector, consisting of all zeros but the $i$-th component that is equal to one.
    \item Constant $g$ is the norm of the gravitational acceleration.
    \item $I_{n \times n} \in \mathbb{R}^{n \times n}$ denotes the identity matrix of size $n$.
    \item  $\prescript{\mathcal{A}}{}{p}_{\mathcal{B}} \in \mathbb{R}^3$ is the the position of the origin of the frame $\mathcal{B}$ with respect to the frame $\mathcal{A}$.
    \item  $\prescript{\mathcal{A}}{}{R}_{\mathcal{B}} \in SO(3)$ represents the rotation matrix of the frames $\mathcal{B}$ with respect to $\mathcal{A}$.
    \item $\prescript{\mathcal{A}}{}{\omega}_{\mathcal{B}} \in \mathbb{R}^3$ is the angular velocity of the frame $\mathcal{B}$ with respect to $\mathcal{A}$, expressed in $\mathcal{A}$.
    \item The operator $\text{sk}(.) :\mathbb{R}^{3 \times 3} \to so(3)$ denotes \textit{skew-symmetric} operation of a matrix, such that given $A \in \mathbb{R}^{3 \times 3}$, it is defined as $\text{sk}(A) := (A - A^\top)/2$.
    \item The operator $S(.) :\mathbb{R}^{3} \to so(3)$ denotes \textit{skew-symmetric} vector operation, such that given two vectors $v,u \in \mathbb{R}^{3}$, it is defined as $v \times u = S(v)u$.
    \item The \textit{vee} operator $.^{\vee} : so(3) \to \mathbb{R}^{3}$ denotes the inverse of \textit{skew-symmetric} vector operator. Given a matrix $A \in so(3)$ and a vector $u \in \mathbb{R}^{3}$, it is defined as $Au = A^{\vee} \times u$.
    \item $_AX^{B}$ denotes a wrench 6D vector transform, as defined in \cite{traversaro2016multibody}, such that $_AX^{B}=\begin{bmatrix} {}^{A}R_B & 0 \\ S(p_B - p_A) & {}^{A}R_B \end{bmatrix}$
    \item The operator $\left\lVert . \right\rVert_2$ indicates vector squared norm. Given $v \in \mathbb{R}^{n}$, it is defined as $\left\lVert v \right\rVert_2 = \sqrt{v_1^2+...+v_n^2}$.
\end{itemize}

\subsection{Modelling}
\label{sec:background:modelling}
A humanoid robot can be modelled as a multi-body mechanical system composed of $n + 1$ rigid bodies, called \textit{links}, connected by $n$ \textit{joints} with one {degree of freedom} (DoF) each. The system is thus assumed to be {floating base}, i.e. none of the links has an {a priori} constant \emph{position-and-orientation} -- hereafter referred to as \emph{pose} --  with respect to the inertial frame $\mathcal{I}$. Hence, a specific frame, attached to a link of the system, is referred to as the \textit{base frame}, and denoted by $\mathcal{B}$.
The \textit{model configuration} is characterized by the pose of the \textit{base frame} along with the \textit{joint positions}. The configuration space lies on the Lie group $\mathbb{Q}=\mathbb{R}^{3+n} \times SO(3) $. An element of the configuration space $q \in \mathbb{Q}$ is defined as the triplet $q = (\prescript{\mathcal{I}}{}{p}_{\mathcal{B}}, \prescript{\mathcal{I}}{}{R}_{\mathcal{B}}, s)$ where $\prescript{\mathcal{I}}{}{p}_{\mathcal{B}} \in \mathbb{R}^3$ and $\prescript{\mathcal{I}}{}{R}_{\mathcal{B}} \in SO(3)$ denote the position and the orientation of the \textit{base frame} respectively, and $s \in \mathbb{R}^n$ is the joints configuration representing the topology of the mechanical system. The position and orientation of a frame $\mathcal{A}$ attached to the model can be obtained via geometrical forward kinematic map $h_{\mathcal{A}}(\cdot):\mathbb{Q} \to SO(3) \times \mathbb{R}^3$ from the \textit{model configuration}.
The \textit{model velocity} is characterized by the linear and angular velocity of the \textit{base frame} along with the \textit{joint velocities}. The  velocity space lies on the set $\mathbb{V} = \mathbb{R}^{6+n}$. An element of the configuration velocity space $\nu \in \mathbb{V}$ is defined as $\nu = (\prescript{\mathcal{I}}{}{\mathrm{v}}_{\mathcal{B}}, \dot{s})$ where $\prescript{\mathcal{I}}{}{\mathrm{v}}_{\mathcal{B}}=(\prescript{\mathcal{I}}{}{\dot{p}}_{\mathcal{B}}, \prescript{\mathcal{I}}{}{\omega}_{\mathcal{B}}) \in \mathbb{R}^6$ denotes the linear and angular velocity of the \textit{base frame}, and $\dot{s}$ denotes the joint velocities. The velocity of a frame $\mathcal{A}$ attached to the model is denoted by $\prescript{\mathcal{I}}{}{\mathrm{v}}_{\mathcal{A}}=(\prescript{\mathcal{I}}{}{\dot{p}}_{\mathcal{A}}, \prescript{\mathcal{I}}{}{\omega}_{\mathcal{A}})$ with the linear and the angular velocity components respectively. The mapping between the frame velocity $\prescript{\mathcal{I}}{}{\mathrm{v}}_{\mathcal{A}}$ and the system velocity $\nu$ is obtained via the \textit{Jacobian} ${J}_{\mathcal{A}}={J}_{\mathcal{A}}(q) \in \mathbb{R}^{6 \times (n+6)}$, i.e. $\prescript{\mathcal{I}}{}{\mathrm{v}}_{\mathcal{A}} = {J}_{\mathcal{A}}(q)  \nu$. The \textit{Jacobian} is composed of the linear part ${J}_{\mathcal{A}}^{\ell}(q)$ and the angular part ${J}_{\mathcal{A}}^{a}(q)$ that maps the linear and the angular velocities, i.e. $ \prescript{\mathcal{I}}{}{\dot{p}}_{\mathcal{A}} = {J}_{\mathcal{A}}^{\ell}(q)  \nu$ and $\prescript{\mathcal{I}}{}{\omega}_{\mathcal{A}} = {J}_{\mathcal{A}}^{a}(q)  \nu$.

\subsection{System Dynamics during Physical Interaction}
\label{sec:background:systemdynamics}
\begin{figure}[t]
	\centering	\includegraphics[trim=0.0cm 0.0cm 0.0cm 0.0cm, clip=true, width=0.75\columnwidth]{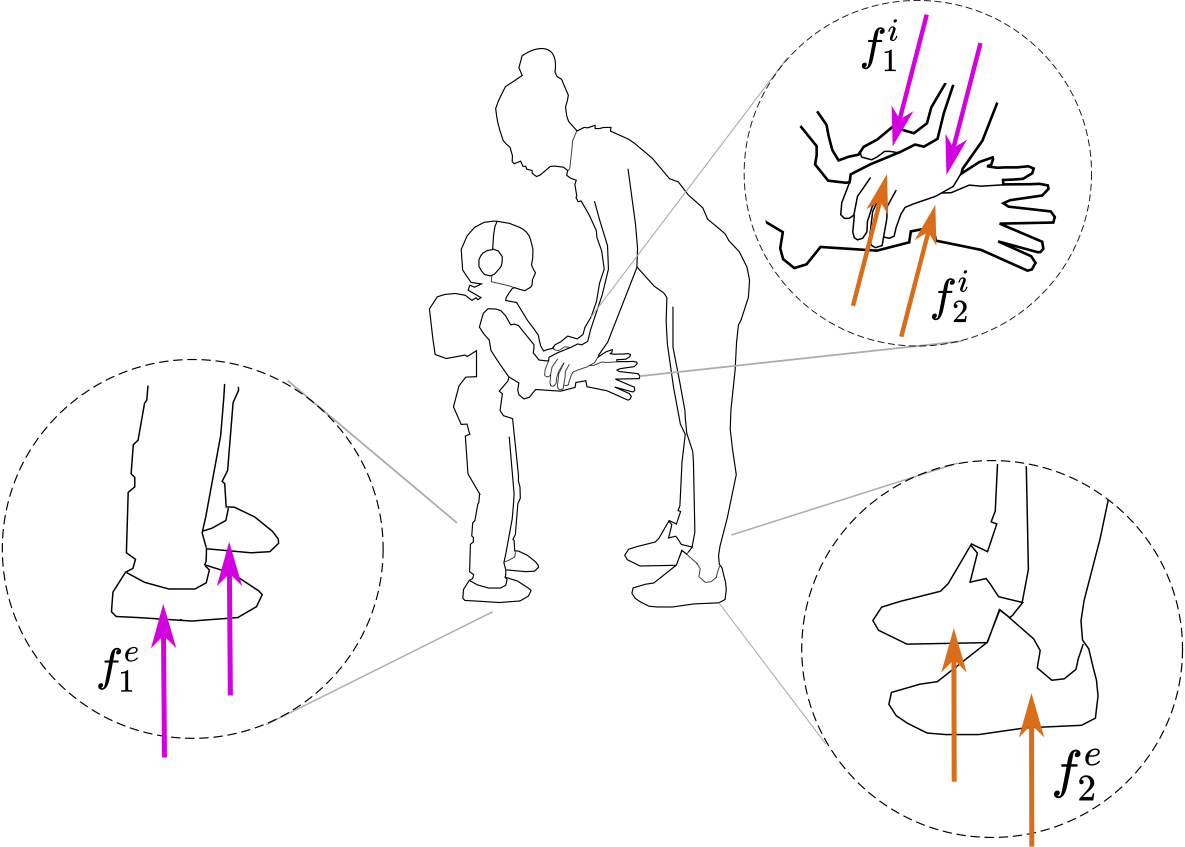}
	\caption{Agent-robot interaction scenario, highlighting the wrenches exchanged with the environment and between the two agents.}
	\label{fig:pHRI}
\end{figure}

The model of each agent is obtained by applying the Euler-Poincarè formalism \cite{Marsden2010}. Agent dynamics is thus described by a set of differential equations complemented with holonomic constraints characterising the contacts:
\begin{align}
\label{eq:constrained-dynamic}
\begin{split}
    & M(q) \dot{\nu} + h(q,\nu) = B {\tau} + J_c^T f, \\
    & J_c \ \nu = 0,
\end{split}
\end{align}
where $M \in \mathbb{R}^{n+6 \times n+6}$ is the mass matrix, the term $h \in \mathbb{R}^{n+6}$ takes into account of Coriolis and gravity forces, $B = (0_{n \times 6},I_n)^T$ is a selector matrix, ${\tau}  \in \mathbb{R}^{n}$ is a vector representing the robot's joint torques, $f \in \mathbb{R}^{6n_c}$ represents the wrenches acting on $n_c$ contact links of the robot, and $J_c \in \mathbb{R}^{n+6 \times 6n_c}$ is the contact Jacobian. 
Now, consider the case of two agents physically interacting, as shown in Fig. \ref{fig:pHRI}. The contacts are not limited to those with the environment, but agent-agent contacts should also be considered. Denoting the terms related to each agent using subscripts, and denoting composite matrices with $\mathbf{bold}$ font, the coupled system dynamics is described  by the following equations \cite{tirupachuri2019}:
\begin{equation}
\begin{split}
\label{eq:multi-system-equations}
& \begin{bmatrix} M_1 & 0 \\\ 0 & M_2\end{bmatrix} \begin{bmatrix} \dot{\nu}_1 \\\ \dot{\nu}_2 \end{bmatrix} + \begin{bmatrix} h_1 \\\ h_2 \end{bmatrix} = \begin{bmatrix} B_1 & 0 \\\ 0 & B_2\end{bmatrix}  \begin{bmatrix} \tau_1 \\\ \tau_2 \end{bmatrix} + \mathbf{Q}^T \mathbf{f}, \\
& \mathbf{Q} \begin{bmatrix} \nu_1 \\\ \nu_2 \end{bmatrix} = 0,
\end{split}
\end{equation}
where $\mathbf{Q}$ is a coupling matrix taking into account both the constraints of the contacts with the environment ($J^{e}\nu=0$) and those of the agent-agent contact points ($J_1^{i}\nu_1=J_2^{i}\nu_2$), and $\mathbf{f}$ is a vector containing all the interaction wrenches (exchanged with the environment and between the two agents) taking into account the action-reaction property for internal forces ($f^{i}_1=-f^{i}_2$) and reflecting the ordering in the constraints matrix $\mathbf{Q}$. Analogously, modelling agents and objects as rigid body, any number of them can be included in the set of equations. Defining properly the composite matrices $\mathbf{M}$, $\boldsymbol{\nu}$, $\mathbf{h}$, $\mathbf{B}$, and $\boldsymbol{\tau}$, any system of multi-rigid bodies with those characteristics can be written compactly as:
\begin{equation}
\label{eq:multi-system-equations-compact}
\begin{split}
& \mathbf{M} \dot{\boldsymbol{\nu}} + \mathbf{h} = \mathbf{B} \boldsymbol{\tau} + \mathbf{Q}^T \mathbf{f} \\
& \mathbf{Q} \boldsymbol{\nu} = 0.
\end{split}
\end{equation}


\section{Force and Postural Ergonomics}
\label{section:ergonomy}

Ergonomics is the scientific discipline concerned with the analysis of human interactions with other elements of the environment, and the design of theory and methods that optimise human well-being and overall system performance~\cite{dul2012strategy}.
When humans interact with robots, optimal ergonomic interactions shall minimize metrics that consider both the human and the robot (here called \textit{agents}). 
In this case, energy expenditure fits well both agents. Human energy expenditure is in fact used to evaluate ergonomics in workplaces \cite{battini2016ergonomics}, and it was considered in many studies aiming at ergonomic human-robot interactions~\cite{kim2018,marin2018,derspaa2020}. Robot energy expenditure is directly related to joint torques, which depend on contact forces and joint posture configuration -- see Eq.~\eqref{eq:constrained-dynamic}. Hence, when choosing joint torques as metric, ergonomy optimization can be decomposed into two sub-problems here referred to as \textit{force} and \textit{postural ergonomy}.

\begin{figure}[b]
\centering
\begin{subfigure}[b]{0.23\textwidth}
  \includegraphics[trim=3cm 17.5cm 5.5cm 2cm, clip=true,width=0.98\columnwidth]{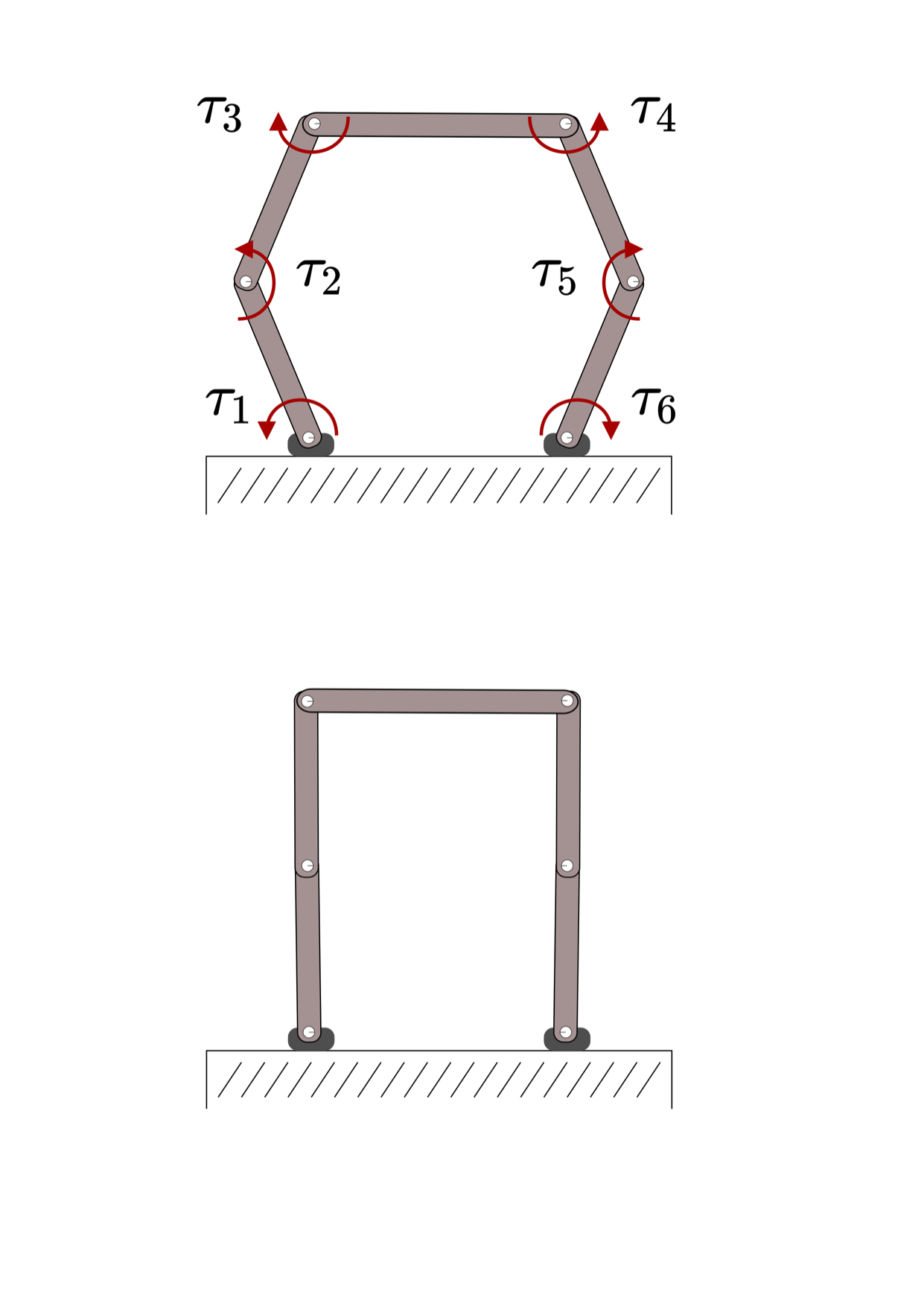}
  \caption{}
  \label{fig:hyperstatic:force}
\end{subfigure}
\begin{subfigure}[b]{0.23\textwidth}
  \includegraphics[trim=3.5cm 4cm 5cm 15cm, clip=true,width=0.98\columnwidth]{figs/ok_final_ergonomyforceposture.png}
  \caption{}
  \label{fig:hyperstatic:posture}
\end{subfigure}
\caption{An hyperstatic mechanical structure with two different configurations: non-optimized (a) and optimized posture (b). 
}
\label{fig:hyperstatic}
\end{figure}

To provide the reader with a better comprehension of the differences between these two metrics, consider the hyperstatic structure shown in Fig.~\ref{fig:hyperstatic}, which exemplifies more complex human-robot scenarios -- e.g. that of Fig.~\ref{fig:pHRI}. This structure is statically indeterminate, so given an equilibrium configuration, e.g. that of Fig.~\ref{fig:hyperstatic:force}, joint torques are not uniquely determined~\cite{matheson1959hyperstatic}. It follows that, despite deliberately high torques, the equilibrium can be maintained when redundant torques compensate each other, resulting in high internal and reaction forces. In this sense, torque minimization can be attempted through force optimization. So, given a system configuration (e.g. a specific set of joint angles), \textit{force ergonomics} aims at minimizing the joint torques at that configuration, eventually through the contact forces.
Now, consider the configuration in Fig.~\ref{fig:hyperstatic:posture}. In this case, the equilibrium is maintained in absence of internal joint torques, showing that an optimized posture can further minimize joint torques. So, \textit{postural ergonomics} exploits the system configuration (e.g. the robot joint angles) at the equilibrium  to further optimize the overall joint torques. 

Note that \emph{postural ergonomics} does not in general imply \emph{force ergonomics}. For instance, consider again Fig.~\ref{fig:hyperstatic:posture}: if the torques 
$\tau_3$ and $\tau_4$ have equal intensity, then the system keeps the equilibrium with unnecessary energy expenditure. 
The shared control architecture presented next 
aims at solving both force and postural ergonomics for a robot-robot interaction scenario. The detailed formulation 
is given in Secs.~\ref{sec:background:stack-of-task} and~\ref{sec:method:posture-optimization}. Let us remark that the \emph{force} and \emph{torque ergonomics} metrics can be used  also to characterise the energy expenditure of a human being interacting with a robot.

\section{Shared Control Architecture}
\label{section:method}
\begin{figure*}[!t]
	\centering
	\includegraphics[trim=0cm 0cm 0 0cm, clip=true, width=0.7\textwidth]{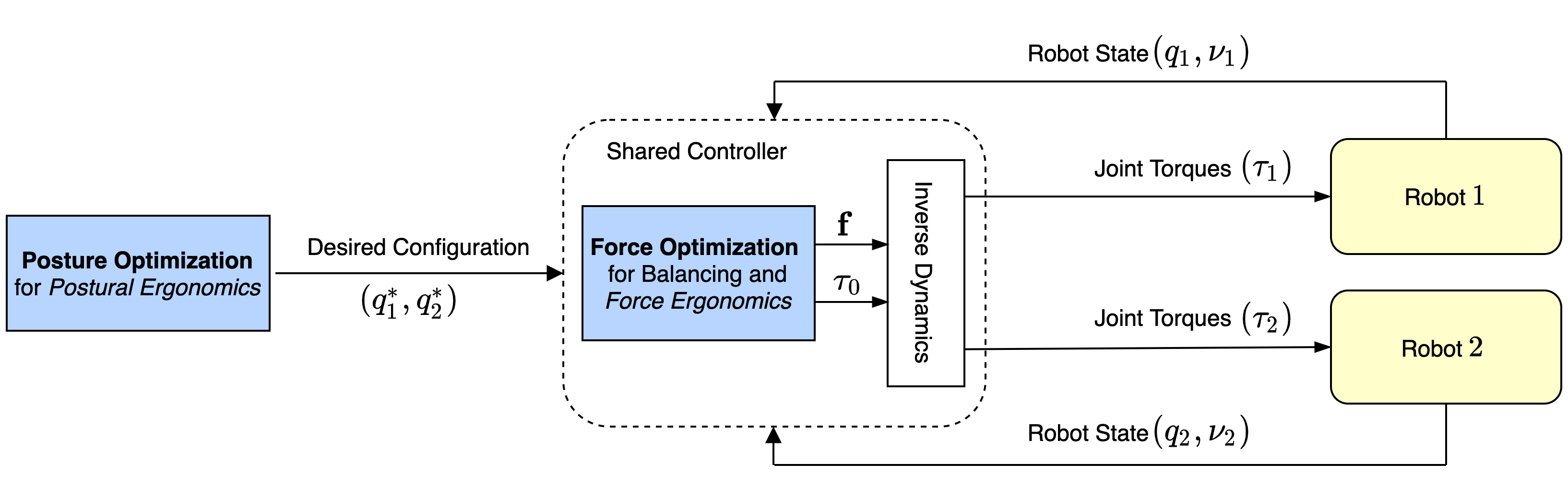}
	\caption{Overview of the shared control framework for robot-robot collaboration with postural and force ergonomic optimization. }
	\label{fig:scheme}
\end{figure*}

The proposed architecture for robot-robot collaborative payload manipulation is shown in Fig. \ref{fig:scheme}.
A shared controller solves the force optimization online taking into account force ergonomics, system balancing, and payload pose control. The optimized contact forces are then used together with a postural task to compute joints torque via inverse dynamics, and send references to the robots.
The reference trajectories of the postural task are computed offline using a second optimization framework that tackles postural ergonomics.

\subsection{System Balancing}
\label{sec:background:total-momentum}
When considering an underactuated floating-base system, the system balancing is a key-factor for the successful achievement of desired task. As mentioned in Sec. \ref{section:introduction}, momentum-based balancing control strategies can ensure a stable robot behaviour against external disturbances.
In a multi-agent scenario, a momentum-based controller can be implemented for each $j$-th robot independently defining
\begin{IEEEeqnarray}{RCL}
\label{eq:signle-robot-momentum-controller}
\IEEEyesnumber
&\dot{H}_j(f_j)=_{G_j}X^{j}f_j+m_jge_3~, \IEEEyessubnumber \label{eq:signle-robot-momentum-controller:dynamics}\\
&\dot{{H}}_j^*:=\dot{{H}}_j^d-K^j_p({H}_j-{H}_j^d)~, \IEEEyessubnumber \label{eq:signle-robot-momentum-controller:feedback-linearization}
\end{IEEEeqnarray}
where $m_j$ is the robot mass, $f_j$ is a vector containing all the wrenches applied to the robot, ($\dot{H}^d_j,H_j^d$) is the reference trajectory for the robot momentum, $K_p^j \in \mathbb{R}^{6\times6}$ is a symmetric positive definite matrix, and $G_j$ is the \textit{center of mass (CoM)} of the robot. Eq. \eqref{eq:signle-robot-momentum-controller:dynamics} describes the dynamics of the robot momentum, while Eq. \eqref{eq:signle-robot-momentum-controller:feedback-linearization} defines the desired dynamics with proportional compensation. Note that $f_j$ takes into account both the wrenches exchanged with the environment $f_j^e$ and among the agents $f_j^m$.
Independent controllers, however, do not exploit the presence of a partner agent. In order to achieve an optimized solution for the entire system, we add an extra task for controlling the \textit{total momentum} $H_T$ of the multi-agent system. The dynamics of total momentum depends on external wrenches only and it is described as,
\begin{equation}
\label{eq:total-momentum-dynamics-external}
    \dot{{H}}_T({f}^e)=_{G}{X}^{e}\mathbf{f}^e+m_Tge_3~,
\end{equation}
where ${m}_T{=}\sum_{j=1}^{N}m_j$ is the total mass of the system, and $\mathbf{f}^e$ the external wrenches vector.
Similarly to Eq. \eqref{eq:signle-robot-momentum-controller:feedback-linearization}, given a desired trajectory for the total momentum ($H^d_T,\dot{H}^d_T$), we define desired total momentum \emph{starred} dynamics.
Same task priority is assigned to robots and total momentum control. Using the representation introduced in Sec. \ref{sec:background:modelling}, the momentum control tasks are written compactly as:
\begin{IEEEeqnarray}{RCL}
\IEEEyesnumber
\dot{\mathbf{H}}(\mathbf{f})=\mathbf{X}\mathbf{f}+\mathbf{m}ge_3, \label{eq:momentum-dynamics-and-feedback-all} \IEEEyessubnumber \\
\dot{\mathbf{H}}^*:=\dot{\mathbf{H}}^d-K_p(\mathbf{H}-\mathbf{H}^d). \IEEEyessubnumber 
\end{IEEEeqnarray}
where $\mathbf{H}=\begin{bmatrix}H_1 & H_2 & H_T\end{bmatrix}^T$ and $\mathbf{X}$ transforms the wrenches into respective CoM location. Notice that the shape of Eq. \eqref{eq:momentum-dynamics-and-feedback-all} is analogous to robot momentum control formulation presented in Eq. \eqref{eq:signle-robot-momentum-controller}.

\subsection{Payload Pose Control}
\label{sec:background:object-pose}
The payload is a rigid body object assumed to be attached to the agents via $n^m$ contact constraints. Given a desired trajectory for the payload position ($p^d, \dot{p}^d, \ddot{p}^d$) and orientation ($R^d,\omega^d, \dot{\omega}^d$), according to feedback linearization control of rigid body pose \cite{olfati2001nonlinear}, the desired body velocity is chosen as
\begin{equation}
    \dot{\mathrm{v}}_{\ell}^* :=\begin{bmatrix} \ddot{p}^d \\\ \dot{\omega}^d \end{bmatrix}-K_d \begin{bmatrix} \dot{p}-\dot{p}^d \\\ \omega - \omega^d \end{bmatrix} - K_p \begin{bmatrix}  p-p^d \\\ \text{sk}(R{R_d}^T)^{\vee} \end{bmatrix},
\end{equation}
where $K_d$ and $K_p$ are positive definite diagonal matrices. Given $\dot{\mathrm{v}}_{\ell}^{*}$, the internal wrenches (also referred to as \textit{grasp forces}) have to satisfy the rigid body dynamics equation
\begin{equation}
     M_{\ell}\dot{\mathrm{v}}_{\ell}^*={}_{\ell}\mathbf{W}^i\mathbf{f}^i+m_{\ell}ge_3,
    \label{eq:object-force-control}
\end{equation}
where $M_{\ell}$ is the payload mass matrix, $m_{\ell}$ its mass, and $_{\ell}\mathbf{W}^i$ is the grasp matrix that transforms wrenches into payload CoM, assuming rigid contacts model \cite{bicchi2001robotic}.
The desired internal wrench solution $\mathbf{f}^i$, obtained from Eq. \eqref{eq:object-force-control}, depends the grasp matrix and on the associated inequality constraints, e.g. frictional contacts model \cite{han2000grasp}, and can be not unique if $\text{rank}(_{\ell}\mathbf{W}^i)>6$. This condition is often verified in presence of multiple contacts. The null-space component of the solution represents the \textit{squeeze wrench}, i.e. the internal wrench that causes no motion of the object.

\subsection{Force Optimization}
\label{sec:background:stack-of-task}
Systems balancing and payload pose control are achieved by regulating the external contact wrenches $\mathbf{f}^e$, and internal wrenches $\mathbf{f}^i$, according to Eqs. \eqref{eq:momentum-dynamics-and-feedback-all} and \eqref{eq:object-force-control}. The solution for those equations can be redundant, therefore, wrenches can be chosen according to force ergonomy optimization. As discussed in Sec. \ref{section:ergonomy}, ergonomics is achieved via joint torque minimization. 
Combining the system dynamics description in Eq. \eqref{eq:multi-system-equations-compact}, the control laws in Eqs. \eqref{eq:momentum-dynamics-and-feedback-all} and \eqref{eq:object-force-control}, and the ergonomy optimization, we have designed an optimization-based shared controller. In the language of optimization it is formulated as,
\begin{IEEEeqnarray}{RCL}
\label{opt:stack-of-task-optimization}
\IEEEyesnumber
\mathbf{f} ^{*} & = & \underset{\mathbf{f}}{\text{argmin}}\left\lVert K_{\tau}\boldsymbol{\tau}(\mathbf{f}) \right\rVert_2 \IEEEyessubnumber\\
& s.t. & \nonumber\\ 
&& C\mathbf{f}<b \IEEEyessubnumber \\
&& \dot{\mathbf{H}}(\mathbf{f}) =\dot{\mathbf{H}}^{*}  \IEEEyessubnumber \\
&& \dot{\mathrm{v}}_{\ell}(\mathbf{f}^i) =\dot{\mathrm{v}}_{\ell}^{*} \IEEEyessubnumber \\
&& \boldsymbol{\tau}^*(\mathbf{f}) = \underset{\boldsymbol{\tau}}{\text{argmin}}\left\lVert \boldsymbol{\tau}(\mathbf{f}) - \boldsymbol{\tau}_0 \right\rVert_2 \IEEEyessubnumber \label{opt:stack-of-task-optimization:joint-torques}\\
&& \quad s.t. \nonumber \\
&& \quad\quad \mathbf{M} \dot{\boldsymbol{\nu}} + \mathbf{h} = \mathbf{B} \boldsymbol{\tau} + \mathbf{Q}^T \mathbf{f} \IEEEyessubnumber \\
&& \quad\quad \dot{\mathbf{Q}} \boldsymbol{\nu} + \mathbf{Q} \dot{\boldsymbol{\nu}} = 0. \IEEEyessubnumber \\
&& \quad\quad \boldsymbol{\tau}_0 = \mathbf{h}_s- \mathbf{Q}^T_s\mathbf{f}-\mathbf{u}_0 \IEEEyessubnumber
\yesnumber
\end{IEEEeqnarray}
where $C$ and $b$ define an additional constraint that ensure the contact and grasping wrenches belong to the associated friction cones, and $\mathbf{h}_s$ and $\mathbf{Q}_s$ are the sub-matrices of $\mathbf{h}$ and $\mathbf{Q}$ relative to joint dynamics only. The stack-of-task optimization in \eqref{opt:stack-of-task-optimization} can be solved in two steps: optimization \eqref{opt:stack-of-task-optimization:joint-torques} is solved as minimum-norm solution obtaining a linear expression for $\boldsymbol{\tau}^{*}(\mathbf{f})$, and optimal wrenches $\mathbf{f}^{*}$ are computed via \textit{Quadratic Programming}. The free variable $\boldsymbol{\tau}_0$ is used to ensure the stability of the zero dynamics by means of the so called \textit{postural task} that, given a desired joint configuration $(s^d_i,\dot{s}^d_i)$, is computed independently for each robot as \cite{nava2016stability}
\[
u_0^i=-K_p^s(s_i-s_i^d)-K_d^s(\dot{s}_i-\dot{s}_i^d).
\] 
\noindent The gain $K_{\boldsymbol{\tau}}$ is a positive definite  matrix used to distributes the effort among the joints and the agents. Assuming matrix 
$K_{\boldsymbol{\tau}}=\begin{bmatrix}k_{\tau_1}I &  \\  & k_{\tau_2}I \end{bmatrix}$ 
is block diagonal, if $k_{\tau_1}=k_{\tau_2}$ the effort is equally distributed among the agents, otherwise, the effort distribution is asymmetric and the ergonomy of agent with higher gain is favoured at the expense of the other agent. Once the optimum value $\mathbf{f}^{*}$  is determined, the robot  torques $\boldsymbol{\tau}$ are obtained by evaluating $\boldsymbol{\tau}^{*}=\boldsymbol{\tau}(\mathbf{f}^{*})$.

\subsection{Postural Optimization}
\label{sec:method:posture-optimization}

\begin{figure}[!b]
\centering
\begin{subfigure}{0.23\textwidth}
  \includegraphics[trim=5cm 0.0cm 5cm 0.0cm, clip=true,width=1.00\columnwidth]{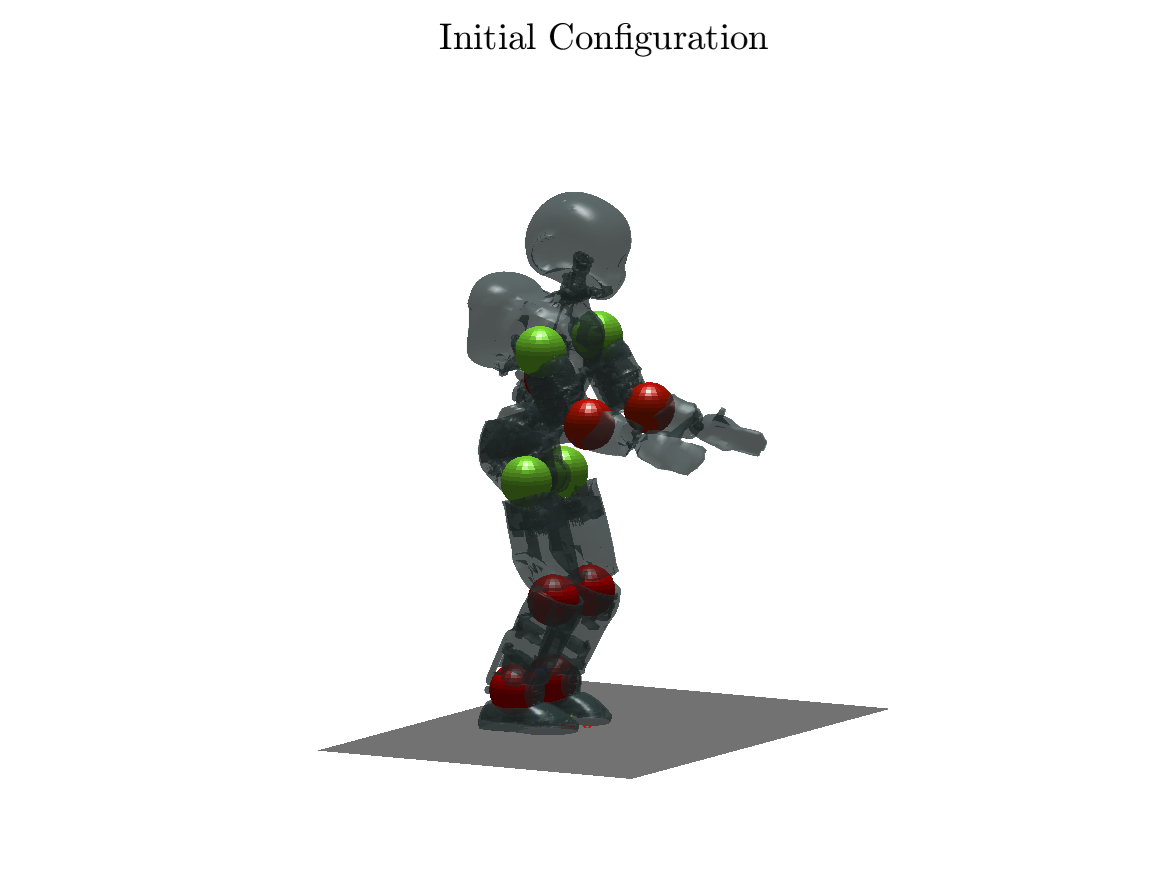}
\end{subfigure}
\begin{subfigure}{0.23\textwidth}
  \includegraphics[trim=5cm 0.0cm 5cm 0.0cm, clip=true,width=1.00\columnwidth]{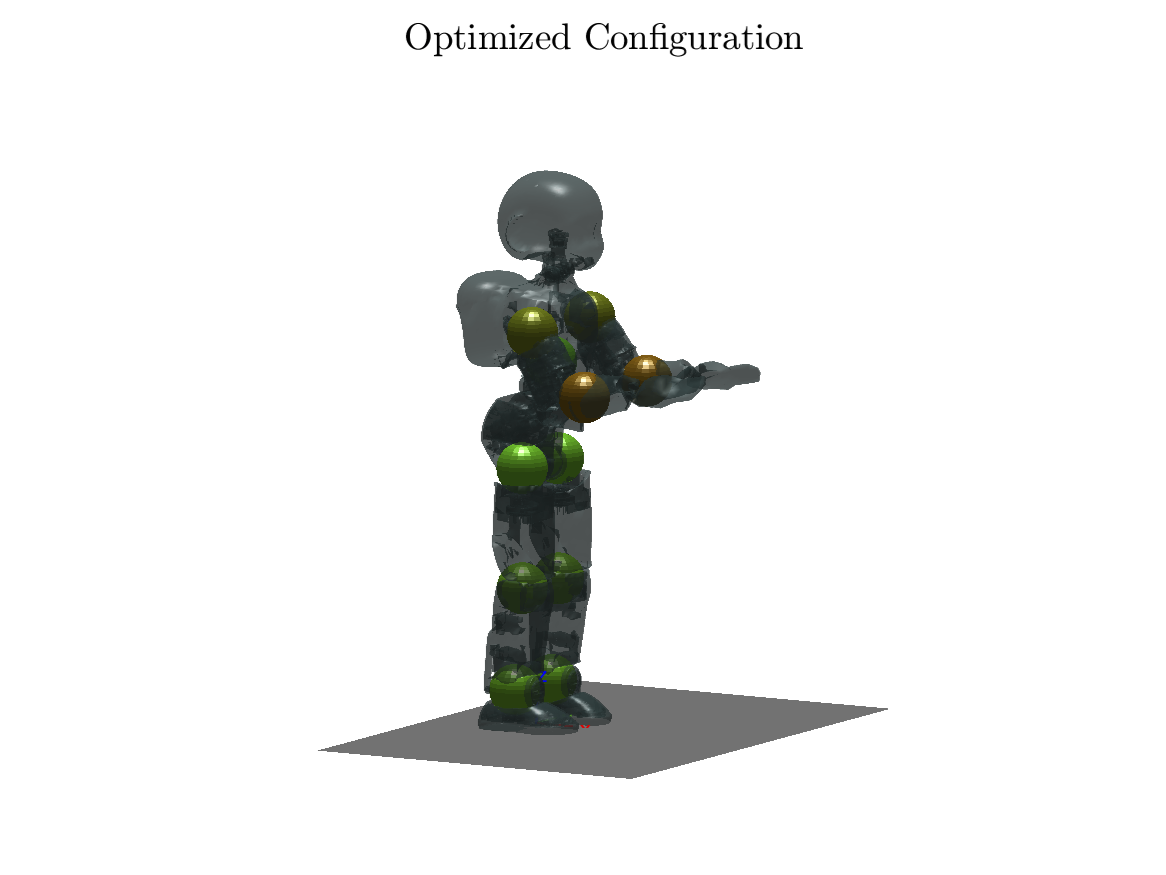}
\end{subfigure}
\caption{On the left, non-optimized robot configuration for collaborative lifting. On the right, optimal configuration minimizing joint torques. The colour of the spheres represents the norm of the joint torques at the given location, ranging from green (low torque) to red (high torque).} 
\label{fig:configuration_optimization}
\end{figure}

The optimization-based controller presented in Sec. \ref{sec:background:stack-of-task} stabilizes the behaviour of the robots during the execution of the task. However, the generation of reference trajectories for momentum tasks  $\mathbf{H}^d$, object pose ($p^d$,$R^d$), and robots configuration $s_j^d$, has not been discussed. The idea is to compute the reference trajectories according to the ergonomy optimization principle of joint torques minimization. Given the geometric properties of the systems, all the above reference quantities can be retrieved from agents configuration, i.e. $\mathbf{H}^d=\mathbf{H}^d(q_1,q_2)$, $p^d=p^d(q_1,q_2)$, and $R^d=R^d(q_1,q_2)$. 
Hence, the search space is reduced to the robots configuration only, and we express the problem as the posture ergonomics optimization discussed in Sec. \ref{section:ergonomy}. 
The optimization problem is formulated as,
\begin{IEEEeqnarray}{RCL}
\label{eq:posture-optimization}
\IEEEyesnumber
(q_1^{*},q_2^{*}) & = & \underset{(q_1,q_2)}{\text{argmin}}\left\lVert K_{\tau}\boldsymbol{\tau}(q_1,q_2) \right\rVert_2 \IEEEyessubnumber \\
& s.t. & \nonumber \\ 
&& C\mathbf{f}<b \IEEEyessubnumber  \\
&& A_js_j<c_j \IEEEyessubnumber  \\
&& \mathbf{M} \dot{\boldsymbol{\nu}} + \mathbf{h} = \mathbf{B} \boldsymbol{\tau} + \mathbf{Q}^T \mathbf{f} \IEEEyessubnumber  \\ 
&& \dot{\mathbf{Q}} \boldsymbol{\nu} + \mathbf{Q} \dot{\boldsymbol{\nu}} = 0 \IEEEyessubnumber \\
&& \boldsymbol{\nu}=0 \IEEEyessubnumber 
\end{IEEEeqnarray}
where $A_j$ and $c_j$ describe the joint limits for the $j$-th robot. As for the shared controller, the optimizer takes into account the presence of the payload trough the coupled dynamics described in Eq. \eqref{eq:multi-system-equations-compact}, and the gain $K_{\tau}$ regulates the distribution of the effort. Differently from the optimization \eqref{opt:stack-of-task-optimization}, however, the problem is non-linear. It follows that the optimization is computationally expensive, and it might be preferred to solve it offline.
The choice of finding the posture at the steady-state, i.e. $\boldsymbol{\nu}=0$, reduces the complexity of the problem, and well approximates the behaviour for slow motions. Given the steady-state optimal configuration, the transition trajectories are generated online as minimum-jerk trajectories, ensuring human-like motion \cite{pattacini2010experimental}. Alternatively, computing dynamic optimization for the whole trajectory is possible, but implies higher computational cost. Fig. \ref{fig:configuration_optimization} presents the initial and optimized configuration for the iCub robot. In the optimized configuration, it can be noticed that singular configurations are preferred, and the effort is better distributed among the joints.


\section{Experimental Results}
\label{section:experiments}

\begin{figure}[t]
	\centering
    \includegraphics[trim=2cm 0cm 3cm 0cm, clip=true, width= 0.85\columnwidth]{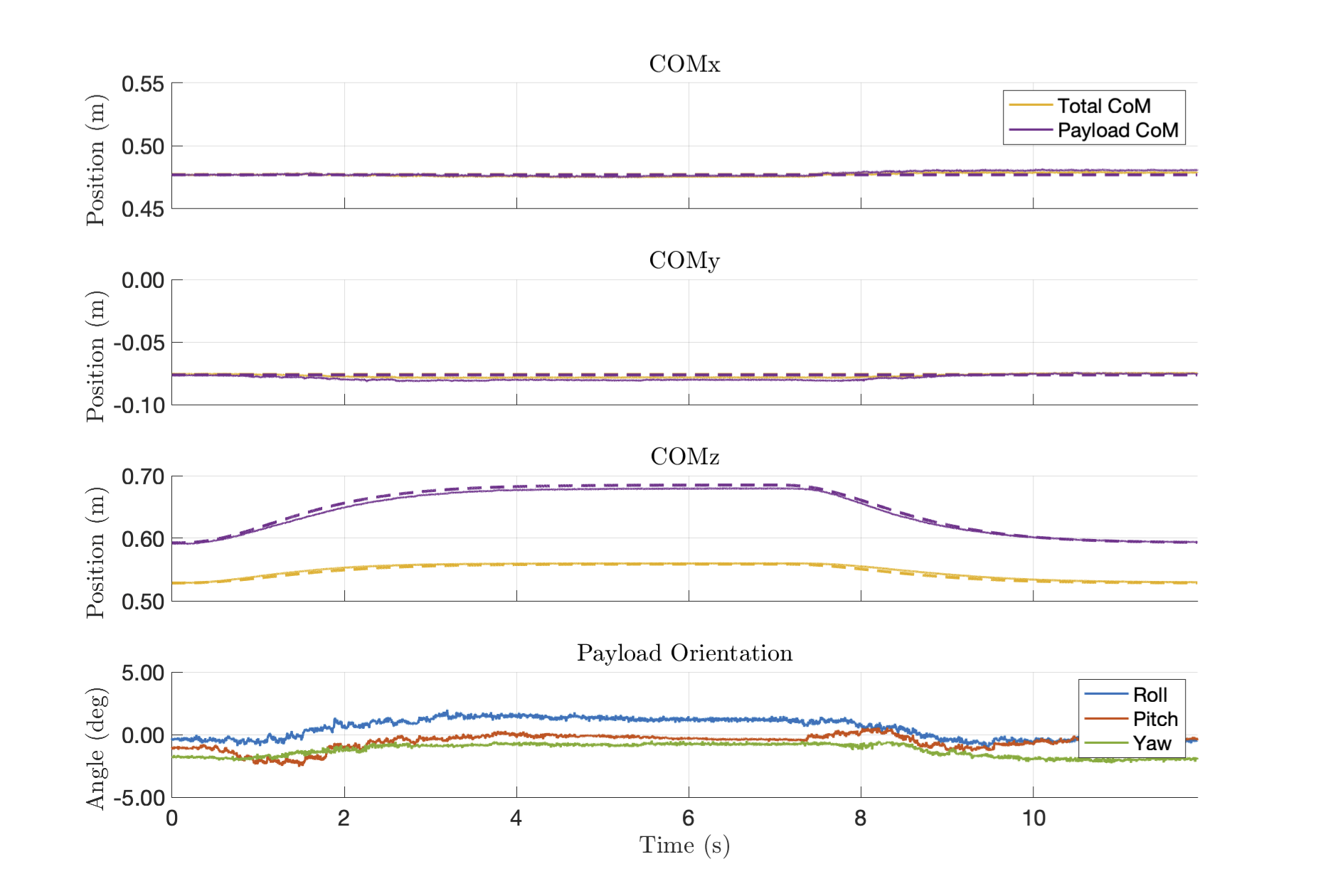}
	\caption{Tracking of the desired trajectories, represented with dashed lines, during collaborative lifting. From the top, the first three rows show the measured position of total and payload centers of mass. The last row shows the payload orientation as roll-pitch-yaw angles.}
	\label{fig:coms-box-tracking}
\end{figure}

The proposed control infrastructure has been tested using two iCub humanoid robots \cite{Natale2017}, endowed with 23 degrees of freedom. According to the robot capability, a payload of 3 Kgs is used, and \textit{ad hoc} mechanical connectors are assembled in order to strength the robot manipulation capability and ensuring rigid coupling with the payload. The robots are required to perform a vertical weight lifting task. Despite belonging to the same major iCub version, the two robots present minor differences in hardware and low-level motor control configuration, therefore, we do not expect to observe perfect symmetric behaviour. The robots communicate with a central server using YARP middleware \cite{fitzpatrick2008towards} trough a Local Area Network. The central server receives robots state information and runs the shared controller at 100 Hz solving the stack-of-task optimization and sending the target joint torques for the robots low-level torque control loops which run at 1 kHz on each robot. The experimental setup is shown in Fig. \ref{fig:real-robots}, all the experiments start from an initial configuration where the payload lies on a support before being lifted by the robots. The non-linear posture optimization problem is solved offline using interior-point method \cite{waltz2006interior}. 
Fig. \ref{fig:coms-box-tracking} shows typical results obtained during collaborative lifting experiments in term of trajectories tracking. The controller appears to be capable of performing balancing while controlling the payload pose. 
The desired vertical displacement of the payload is tracked for the entire duration of the experiments, and it rotates few degrees from the ground plane. Good tracking is also achieved for the total center of mass trajectory.

\begin{figure}[!b]
	\centering
    \includegraphics[trim=1.4cm 0cm 1.6cm 0cm, clip=true, width=0.95 \columnwidth]{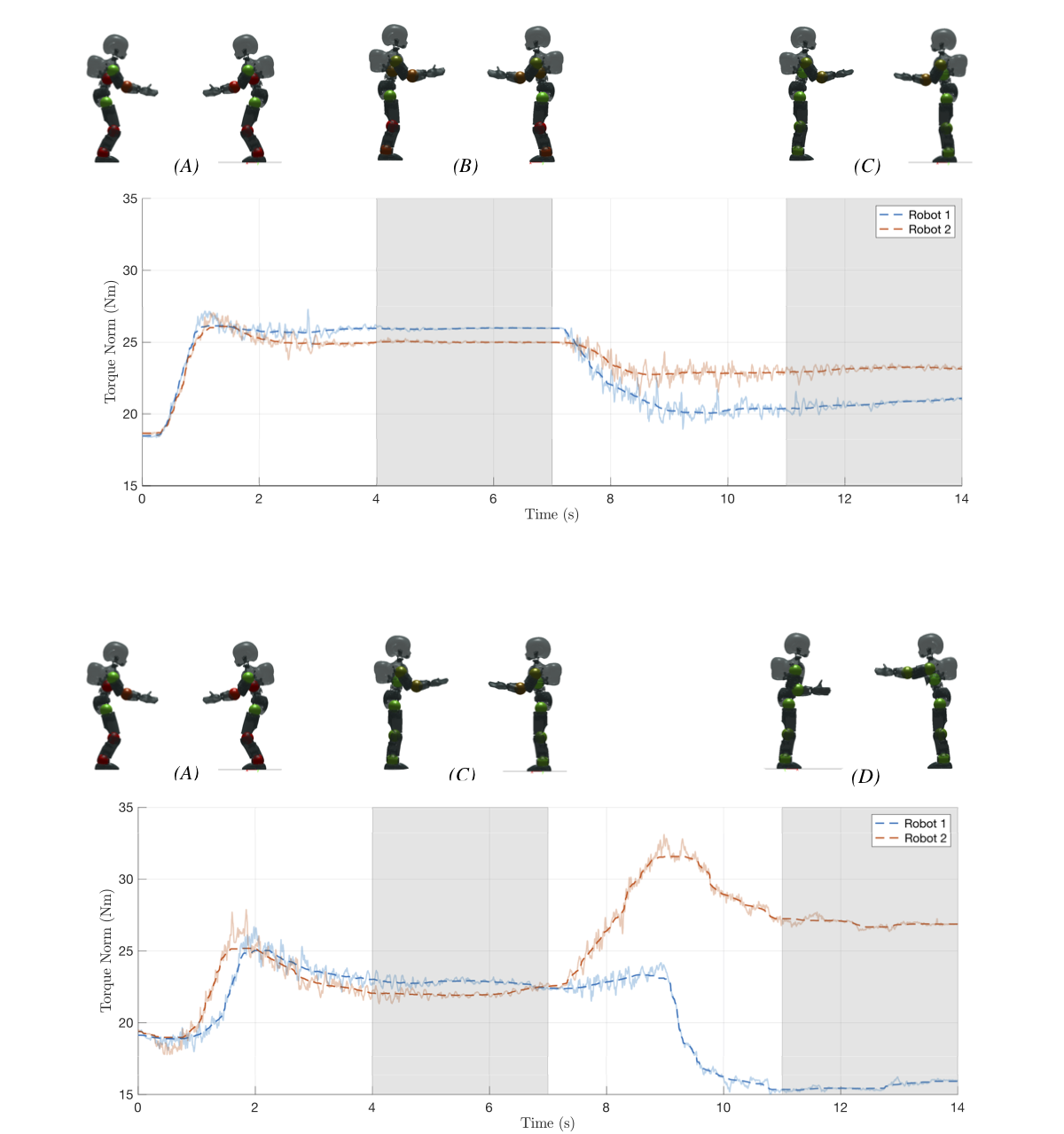}
	\caption{Measured joint torques vector norm during robot-robot collaborative lifting. Dashed lines have been obtained smoothing the original torque measurements. Two sequences are presented, starting from the same initial configuration \textit{(A)}. In the first sequence, the robots first lift the payload \textit{(B)}, and then move to ergonomic configuration \textit{(C)}. In the second sequence, the robots lift the payload directly using ergonomic reference configuration \textit{(C)}, and then move to a configuration with asymmetric effort distribution \textit{(D)}.
	}
	\label{fig:motion-sequences}
\end{figure}

Different motion sequences have been tested in order to evaluate the ergonomy optimization framework through joint torques measurement. In particular, the two sequences presented in Fig. \ref{fig:motion-sequences} are well explicative, and demonstrate the effects of ergonomics optimization. In both cases the robots start from the same initial configuration \textit{(A)}, before moving among three possible configurations: \textit{(B)} is a \textit{non-optimized configuration}, \textit{(C)} is the \textit{optimized ergonomic configuration} obtained from \eqref{eq:posture-optimization} with $k_{\tau_1}=k_{\tau_2}$, and \textit{(D)} is an \textit{asymmetric configuration} computed from \eqref{eq:posture-optimization} using unequal gains such that $k_{\tau_1}<k_{\tau_2}$. When moving from the non-optimized to the optimized posture, the torques vector norm decreases for both the robots showing the effect of ergonomic optimization in achieving a more efficient configuration. Considering the transient time, in the transition \textit{(A)-(C)} the torques norm is lower compared to transition \textit{(A)-(B)}, despite the optimization is solved considering the steady-state only, but it appears to have an higher percentage overshoot.
Analysing the asymmetric configuration \textit{(D)}, the torques norm shows that the framework succeeds in overloading \textit{Robot 2} to accommodate \textit{Robot 1} ergonomics.

\section{Conclusions}
\label{section:conclusioins}
In this paper, a shared control framework for multiple humanoid robots collaboration is proposed. Thanks to a centralized controller, the joint action of the robots is controlled to guarantee the balancing of the system and the control of the payload during a lifting task. Moreover, efficient collaboration is ensured by taking into account the ergonomics requirements of the robots. The contact forces and the robot posture are optimized in order to distribute the effort among the agents minimizing energy consumption. Preliminary tests with the iCub humanoid robot verify the soundness of the proposed framework, but further experiments are required in order to have a statistical validation. 

As future work we plan to extend the architecture to a more complex experimental scenarios involving different type of grasping and agents locomotion.
Another interesting extension can be the application of the proposed algorithm for human-robot collaboration with the objective of optimizing human ergonomics through robot assistive actions. 
Finally, the precise definition of the force and posture ergonomics principles  introduced in Section \ref{section:ergonomy} will be the subject of a forthcoming publication.



\section*{Acknowledgments}
This work is supported by Honda R\&D Co., Ltd and by EU An.Dy Project that received funding from the European Union’s Horizon $2020$ research and innovation programme under grant agreement No. $731540$. The content of this publication is the sole responsibility of the authors. The European Commission or its services cannot be held responsible for any use that may be made of the information it contains.

 
\bibliographystyle{IEEEtran}
\bibliography{bibliography}

\end{document}